# Deep Neural Network Based Ensemble learning Algorithms for the healthcare system (diagnosis of chronic diseases)

Jafar Abdollahi [a], Babak Nouri-Moghaddam [a,b,*], Mehdi Ghazanfari [b]

[a] Department of Computer Engineering Ardabil Branch, Islamic Azad University Ardabil, Iran
[b] Department of Industrial Engineering, Iran University of Science and Technology, Tehran, 1684613114, Iran

Graphical Abstract

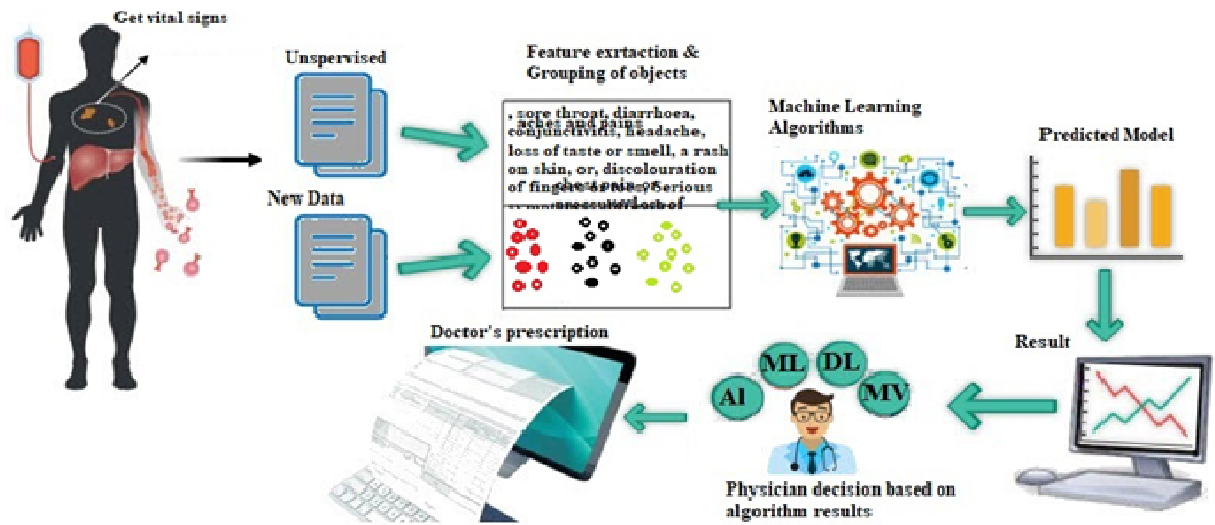

**Abstract**

**Objectives**: Purpose, realistic. Diagnosis of chronic diseases and assistance in medical decisions is based on machine learning algorithms. In this paper, we review the classification algorithms used in the health care system (chronic diseases) and present the neural network-based Ensemble learning method. We briefly describe the commonly used algorithms and describe their critical properties.

**Materials and Methods**: In this study, modern classification algorithms used in healthcare, examine the principles of these methods and guidelines, and to accurately diagnose and predict chronic diseases, superior machine learning algorithms with the neural network-based ensemble learning Is used. To do this, we use experimental data, real data on chronic patients (diabetes, heart, cancer) available on the UCI site.

**Results**: We found that group algorithms designed to diagnose chronic diseases can be more effective than baseline algorithms. It also identifies several challenges to further advancing the classification of machine learning in the diagnosis of chronic diseases.

**Conclusion**: The results show the high performance of the neural network-based Ensemble learning approach for the diagnosis and prediction of chronic diseases, which in this study reached 98.5, 99, and 100% accuracy, respectively.

## 1. INTRODUCTION

Their several programs for machine learning (ML), the most important of which is the accurate prediction of chronic diseases. Each instance in each data set used by machine learning algorithms is represented using the same feature set. Attributes may be continuous, batch, or binary. If samples are given with known labels (corresponding correct outputs), it is called supervised learning, as opposed to unsupervised learning where samples are unlabeled [1]. However, for health care epidemiologists to make optimal use of this data, computational techniques are needed that can manage large complex data sets. Machine learning (ML), studying tools and methods for identifying patterns in data, can help. The appropriate application of ML in these data are promising as it broadly alters the patient risk classification in the medical field and especially in infectious diseases [2-3-4-5-6-7].

After a better understanding, the strengths and limitations of each method, the possibility of combining two or more algorithms to solve the problem should be considered. The goal is to use the strengths of one method to complement the weaknesses of another. If we are only interested in the best possible classification accuracy, it is difficult or impossible to find a single classification as well as a good set of classifications. Based on a study conducted in 2018, researchers have reached this conclusion. The use of an algorithm in the diagnosis and prediction of chronic disease has not been effective and accurate and has not been successful in many scenarios. This is because, in supervised group methods, they build a set of basic learners (specialists) and use their weighted results to predict new data. Numerous empirical studies confirm that group methods often work better than both individual learners [8-9-10-11-12-13-14].

Classification is one of the most important machine learning techniques. This article summarizes and analyzes the main features of each algorithm by analyzing and comparing the types of conventional classifiers to provide a basis for improving old algorithms or developing new effective ones. This summary can also be used to select these data mining techniques for new applications [15]. In the following, we will review machine learning techniques and review them in general.

1.1. Artificial intelligence techniques

Now we will discuss some fundamental AI techniques: Heuristics, Support Vector Machines, Neural Networks, the Markov Decision Process, and Natural Language Processing. Artificial intelligence techniques used in the process of analyzing complex data and data rich in semantics as well as designing intelligent information systems of the new generation are now dedicated to various areas of use.

1.1.1. Machine Learning:

Machine learning is the scientific study of algorithms and statistical models used by computer systems that benefit from patterns and inferences to perform tasks instead of using explicit instructions. The main focus of machine learning is on the development of computer programs that can access data and use it for their learning. Classification is one of the most important aspects of supervised learning. The goal is to find a model that best describes the data (Fig 1). Usually, all machine learning algorithms are divided into 4 groups based on learning style, performance or problems they solve: supervised learning, unsupervised, semi-supervised, and reinforced learning [16-17-18-19-20].

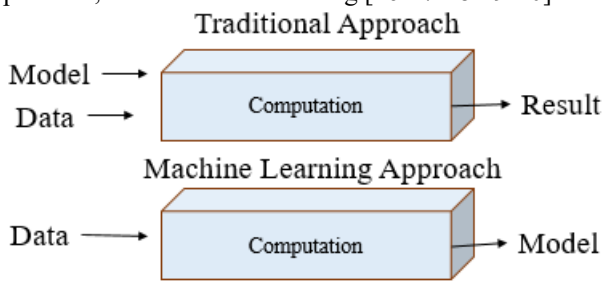

**Fig. 1**. General performance of machine learning

2.1.1. Supervised learning

A "teacher" gives the computer the example inputs and the desired outputs each, and the goal is to learn a general rule that takes the inputs to the output. In certain cases, the input signal may be only partially available or limited to specific feedback. Most machine learning methods use supervised learning. In supervised machine learning, the system tries to learn from the a priori examples provided. In other words, in this type of learning, the system tries to learn the patterns based on the given examples [21-22-23-24]. Mathematically speaking, learning is supervised when the input variable (X) and the output variable (Y) are present and an algorithm can be used to derive an input-to-output mapping function based on them. The mapping function is shown as Y = f (X). Example:

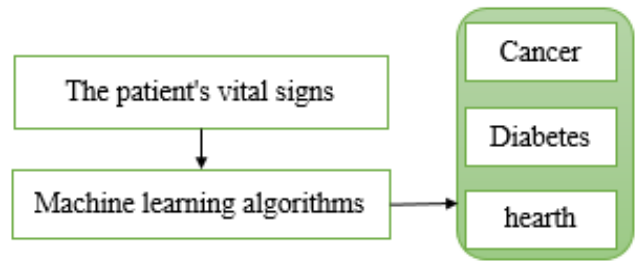

**Fig. 2**. Learning performance with supervision

To train the system, you must provide the system with a large number of samples, or data, including tags and predictors.

3.1.1. Semi- Supervised learning

The computer is given only one incomplete training signal: a training set that some (often many) of its target outputs are missing. It uses unlabeled data and labeled data simultaneously to improve learning accuracy [25-26-27-28-29-30].

4.1.1. Reinforcement learning

Educational data (in the form of rewards or punishments) is given as feedback to program activities only in a dynamic environment, such as driving a car or playing against an opponent [31-32-33-34-35]. Figure 3 shows the structure of the reinforcement learning algorithm.

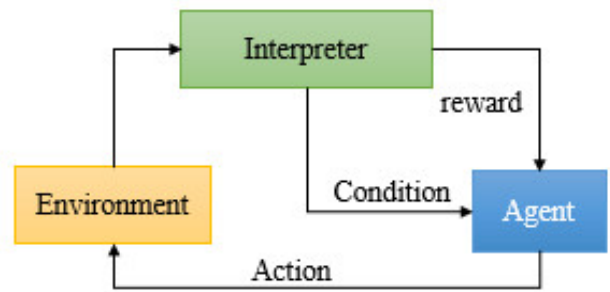

**Fig. 3**. General function of reinforcement learning

5.1.1. Unsupervised learning

No label is given to the learning algorithm and the algorithm itself must find a structure in the input. Self-directed learning can be a goal in itself (finding hidden patterns in data), or a means to an end (learning to show) [36-37-38-39-40-41-42-43]. In fact, without any help from The system you want to recognize that each user belongs to a cluster and it finds connections without your help.

### 6.1.1. Classification
Classification means placing each person, object, and… in different categories under study. Classification helps to analyze unit measurements to identify the category to which they belong. Analysts use data to build an effective relationship. For example, before a bank decides to distribute a loan, it evaluates the ability of the customer to repay the loan. The bank does this by taking into account factors such as customer income, savings, and financial history. This information is obtained from the analysis and classification of past loan data [44-45-46].

### 7.1.1. Prediction
Machine learning can be used in forecasting systems. For example, according to what was said above to provide a bank loan, to calculate the probability of system error, it is necessary to classify the data available in different groups. This set is defined by the rules set by analysts. After classification, the probability of error can be calculated again with the help of machine learning. These calculations can be used in all sectors for a variety of purposes. Prediction is one of the best applications of machine learning [47-48-49].

### 8.1.1. Extraction
One of the great applications of machine learning is information extraction. Here machine learning aids the process of extracting structural information from unstructured data. Machine learning, for example, extracts information from web pages, articles, blogs, business reports, and emails. The database maintains the relationship of the generated output with the information extraction. The machine learning extraction process takes a set of documents as input and extracts structured data from it [50-51].

### 9.1.1. Regression
Machine learning can also be used for regression. In regression, the principle of machine learning can be used to optimize parameters. Machine learning can also be used to reduce the approximate error and calculate the closest possible result. The general form of each type of regression is:

- **Simple linear regression:**
$$Y = a + bX + u \quad (1)$$

- **Multiple linear regression:**

$$Y = a + b_1 X_1 + b_2 X_2 + b_3 X_3 + ... + b_t X_t + u \quad (2)$$

Where:
- Y = the variable that you are trying to predict (dependent variable).
- X = the variable that you are using to predict Y (independent variable).
- a = the intercept.
- b = the slope.
- u = the regression residual.

Machine learning can also be used to optimize performance, and machine learning can be used to change inputs to the closest possible outcome. Machine learning can be used in the techniques and tools used to diagnose diseases. This technology can be used to analyze clinical parameters and combine them to predict disease progression, extract medical information, research to achieve results, treatment planning, and patient monitoring. These are successful applications of machine learning methods. The use of machine learning algorithms can also help integrate computer devices and healthcare departments [51-52-53].

Basic topics in classification algorithms Classifier is an algorithm that plots input data to a specific Ensemble.

- **Classification model**: A classification model tries to deduce from the input values given for training. Predicts class tags/categories for new data.
- **Attribute**: Attribute is an individual measurable attribute of a phenomenon that is observed.
- **Binary Classification**: The task of classifying with two possible outcomes. For example gender classification (female / male)
- **Multi-class Classification**: Classification with more than two classes. In multi-class classification, each instance is assigned to one and only one target tag. For example, an animal can be a cat or a dog, but not both. Multi-tag classification: A classification task in which each instance is drawn into a set of target tags (more than one class). For example, a news article can be about a person, a place, and a place at the same time. [16-53].

Suppose we knew the density, pi(x), for each of the N classes. Then, we would predict using:

$$F(x) = \arg\max_{i \in 1,...,N} pi(x) \quad (3)$$

Of course, we don't know the densities, but we could estimate them using classical techniques.

In this paper, the proposed new methods for diagnosing chronic diseases with the help of artificial intelligence algorithms in several studies (from 2000 to 2020) are reviewed, compared, and implemented. The purpose of reviewing these methods is to help in timely and accurate diagnosis. Clinicians rely on their knowledge and experience, as well as the results of complex and time-consuming clinical trials, to diagnose malignant and dangerous diseases despite the unavoidable human error. The use of machine learning sheds light on the ability of these techniques to help diagnose malignancies.

**The remaining of this article is mainly described as follows.** Section 2 reviews the past work in machine learning for the diagnosis of chronic diseases. The proposed method is introduced in detail in Section 3. Section 4 presents the case study. Finally, the conclusion summaries the paper and proposed future work in Section 5.

## 2. Related Works

Today, the application of artificial intelligence in the field of health systems has greatly expanded. Machine learning, as a sub-branch of artificial intelligence, has many applications in the field of medical diagnosis. Chronic diseases are one of the most common diseases in the world, which facilitating and accelerating its diagnosis will have very favorable results on its future treatment process.
In the studies of Mr. Lam, K. Y et al. Used an activity tracking system and monitoring of Alzheimer's patients using various machine learning algorithms including SVM, RF, NB for position detection, and SmartMInd health care tool for motion detection [54]. To accurately analyze medical data to diagnose primary disease and care for patients, Mr. Chen, M et al. [55] used a neural network algorithm to effectively predict chronic disease. Mr. Babič, F et al. [56] used statistical methods such as decision tree, NB, SVM, and Apriori algorithm to process three sets of medical data to produce prediction models by extracting appropriate rules, which yielded acceptable results. Mr. Paul, A. K et al. [57] proposed a fuzzy system based on a genetic algorithm to predict the risk of heart disease, and the simulation results showed the high performance of the proposed method.

In the studies of Mr. Safdar, S et al. [58] for the diagnosis of heart disease in clinical settings, the machine learning techniques of the ANN algorithm with 97%, Cart algorithm with 87.6% accuracy, logistic regression with 72% accuracy have been achieved. Mr. Hongxu, YI N et al. [59] A hierarchical health decision support system for the diagnosis of a variety of diseases such as arrhythmia (86%), type 2 diabetes (78%), urinary bladder disorders (99%), and kidney stones accurately (94%) and hypothyroidism reached (95%) accurately. In the studies of Mr. Yip, T. F et al. [60] used the development and validation of a machine learning model based on laboratory parameters to diagnose and predict liver disease for the general population. Then, TING-TINGZHAO1 et al [61], used correlation analysis to obtain useful features for classifying diseases using a support vector machine and obtained better results.
In a 2017 study, Zhang, J. et al [62], used neural networks to predict disease based on medical history and used an in-depth learning method to predict the risk of multiple diseases. Mr. Tripoliti, E. E et al. [63] proposed an automated method for early detection of class changes in patients with heart failure using classification techniques with the highest impact factor of 97.87.67%, respectively. Mr. Kim, E. Y et al. [64] used a machine learning method to classify depressive disorder and control groups using proteomic analysis of serum and heart rate changes to identify biomarkers of the new environment. The above performance of the above methods has been to accurately diagnose these disorders.

In a 2017 study, Mr. Lee, S et al. [65] proposed a method for detecting abnormal electronic waveforms based on transplant criteria for effective diagnosis of heart disease. In this experiment, the accuracy of abnormal waveform extraction using real ECG data for a healthy person and a patient with myocardial infarction was evaluated and the results showed that the proposed method can not only sufficiently disrupt the abnormal waveform., Can provide a comprehensive review. Mr. Dominguez-Morales et al. [66] presented a convolutional neural network-based tool for classifying healthy and pathologically ill individuals using the Newomorphic Auditory Sensor for FPGA, which is capable of real-time frequency-band audio decomposition.

From this approach, using different values, the best amount is 97.05% and in the worst case, the accuracy is 80%.
In the studies of Mr. Timothy, V et al. [67], to automatically diagnose various diseases based on heart rate changes, they used machine learning techniques. The process of diagnosing the input data through electrocardiogram recording has reached high accuracy. In the following, Mr. Ekız, S et al. [68] have used six methods of different learning algorithms to classify heart disease in MATLAB and Veka environments, which are: linear SVM, two-way SVM, percussion SVM, intermediate SVM, intermediate SVM Gauss, and the decision tree use and compare these methods.

Also, in a study by YU-BOYUAN1 et al. [69], they used a machine learning method (PSSVM) to help diagnose heart disease. Mr. Kumar, M et al. [70] used the electrocardiogram signal for the automatic diagnosis of coronary artery disease (CAD) and obtained a remarkably accurate classification result for the wavelet nucleus of 99.6%, which is 99.56% compared with the RBF nucleus. Mr. McRae, M. P et al. [71] use a multi-criteria measurement system based on logistic regression techniques to diagnose and predict heart failure and heart health. The experimental results in this paper show the model. scorecard has a high improvement over other cases with an accuracy of AUC = 0.8403 and 0.9412.
Mr. Uçar, M. K. et al. [72] studied the study of sleep-wake determination using KNN and SVM classification algorithms for which the heart rate change signal (HRV) signal was obtained from PPG. Based on the classification results, the classification accuracy is 73.36%, respectively. The sensitivity was 81% and the characteristic was 77%. Also, Mr. Acharya, U. R et al. [73] used coronary artery disease as the main cause of heart attacks and used a reliable and efficient automatic diagnosis system for early detection of CAD. Researchers studied the structures of CNNS (NET1). To identify the classes, the proposed system has an accuracy of 94.95%, a sensitivity of 93.72%, a specificity of 95.18% for the NET1 class, and an accuracy of 95.11%, a sensitivity of 91.13%, and a specificity of 95.88% for the NET2 class. Also, Mr. Kumar, M et al. [74] have used advanced signal processing techniques to diagnose CAD.

Mr. Gagnon, L. L [75] then developed a wireless electrocardiographic recording device that detects and records the ECG system of the heart via an electrode and is connected to a prototype by a host computer via a low-power wireless connection. It is controlled that ECG data can be obtained and displayed in real-time for fast analysis. Mr. Narula, S et al. [76] also studied a machine learning framework that uses electrocardiogram data to automatically evaporate the heart of cardiac patients from the physiological hypothalamus, which developed three machine learning models such as SVM, random forest, and artificial neural network. Mr. Degregory, P. R et al. [77] developed the potential to identify early heart rate markers using a device based on an antibody test, followed by Mr. Medved, et al. [78] Deep learning techniques for modeling in the form of a two-layer neural network to predict waiting time as 1-wait 2-link and 3-dead at three different times of 180.365 and 730 days are presented.

Mr. Luo, Y et al. [79] study an approach based on magnetic resonance imaging (MRI) of the heart, which is a passive data mining imaging technique that uses left ventricular MRI to assess stroke volume, myocardial infarction, and parameters. The performance of areas such as wall movement and wall thickness is very important. In this approach, machine learning methods for analyzing medical images with estimates of more complex models using training data have been considered, which is a proposed fast, robust, efficient, and suitable method for LV division. Mr. Nilashi, M; Et al. [80] have proposed an analytical method for disease prediction using machine learning techniques, from Cart tree regression classification to generate fuzzy rules based on the EM algorithm for data clustering, from the Cart algorithm to rule detection, from The Pca algorithm has used fuzzy rule-based methods for prediction to reduce dimensions and deal with the multipolar problem in data.

Mr. Ali, B et al. [81] presented machine learning techniques for classifying diabetes and cardiovascular disease using artificial neural networks and Bayesian networks. The purpose of this reference is to study artificial neural networks and Bayesian networks and their application in classification. Diabetes and cardiovascular disease are designed to compare machine learning techniques to achieve the highest classification output. Also, Mr. Arabasadi, Z [82] used artificial neural networks and genetic algorithms to effectively diagnose heart disease; Which presents a highly hybrid method for the diagnosis of coronary artery disease and has obtained 93.85%, 97%, and 92% accuracy, sensitivity, and specificity in the Z-Alizadeh Sani dataset. Heart disease has been proposed as a Enseble-based approach that creates new models by combining posterior probabilities or predicted values from multiple previous models to create more effective models that classify 89.01% of tests based on data from the Heart Database database. Cleveland has been extracted and has also reached 95.95% and 91.95% specific sensitivity and specificity in the diagnosis of heart diseases, respectively.

Browse Algorithms: Classification is a science that builds a model for predicting new data labels based on previously labeled data. Classification is one of the basic sub-disciplines of machine learning and data mining, And it is based on data collected from past actions. Actions are labeled based on expert knowledge. To have a good classification model, we need to be aware of the data and its structure, as well as the number of categories (tag-class-class). Although it is sometimes impractical to become familiar with the structure and type of data, the right classification model can sometimes be chosen if there is a simple familiarity. Now that we are familiar with the classification of data mining algorithms, we will introduce the best data mining algorithms.

In the world of data mining, 10 Top Algorithms are very widely used and have high power. These 10 top algorithms are among the most effective data mining algorithms in the research community. Using each algorithm, we provide an algorithm description, implement, compare, and discuss the algorithm's effectiveness in diagnosing chronic diseases, and review current and more research on algorithms.

## 1.2. These 10 algorithms are as follows:

1. Linear regression
2. Logistic regression
3. Naive Bayes Classifier
4. Nearest Neighbors K
5. Decision tree (ID3, C4.5, cart, CHAID, MARS)
6. Random forest
7. Support for vector machines
8. CART algorithm
9. Ensemble learning algorithms (bagging, Boosting, AdaBoost)
10. Neural networks

### 1.1.2. Linear regression

In machine learning, we have a set of input variables (x) that are used to determine the output variable (y). There is a relationship between input variables and output variables. ML's goal is to quantify this relationship. In linear regression, the relationship between input variables (x) and output variables (y) is expressed as the equation of the form $y = a + bx$; therefore, the purpose of linear regression is to discover the values of coefficients a and b. Here, a is intercepted and b is the linear slope [84-85-86-87].

A linear regression line has an equation of the form $Y = a + bX$, where $X$ is the explanatory variable and $Y$ is the dependent variable. The slope of the line is $b$, and $a$ is the intercept (the value of $y$ when $x = 0$).

### 2.1.2. Logistic regression

Logistic regression is a calculation used to predict a binary outcome: either it happens or it does not. This can be displayed as Yes/No, Pass/Fail, Alive/Dead, etc. Independent variables are analyzed to determine the binary result with the results divided into one of two categories. Independent variables can be batch or numeric, but the dependent variable is always the batch. It is written as follows:

$$P(Y-1|X) \text{ or } P(Y=0|X) \qquad (4)$$

Calculate the probability of the dependent variable Y for the independent variable X. This can be used to calculate the probability of a word having a positive or negative meaning (0, 1, or between scales). Or it can be used to determine the object in the photo (tree, flower, grass, etc.), with a probability of 0 to 1 for each object [88-89-90-91]. Mathematically, logistic regression estimates a multiple linear regression function defined as:

$$\log it(p) = \log\left(\frac{p(y=1)}{1-(p=1)}\right) = \beta_0 + \beta_1.x_{i2} + \beta_2.x_{i2} + \cdots + \beta_\varphi.x_a \qquad (5)$$

### 3.1.2. Naive Bayes Classifier

Calculate the possibility of whether a data point is in a specific group or not. In-text analysis, it can be used to classify words or phrases as belonging to a predefined "tag" (classification). To decide whether a phrase should be labeled "sport", you should calculate:

$$P(A|B) + P(AB|A) x P(A))/P(B)) \qquad (6)$$

Or ... probability A, if B is true, is equal to probability B, if A is true, multiples of probability A is true, divisor by probability B is true; so we use Bayes' theorem to calculate the probability of an event occurring, given that another event has already occurred. To calculate the probability that Hypothesis (h) is true, based on our previous knowledge (d), we use the Bayes theorem as follows:

$$P(h|d) = ((P(d|h)P(h)))/(P(d))) \qquad (7)$$

where:

P (h | d) = Posterior probability.
The probability of hypothesis h being true, given the data d, where

$$P(h|d) = P(D1|h)P(d2|h)...p(dn|h)P(d) \qquad (8)$$

$P(d | h) = Likelihood$.
The probability of data d given that the hypothesis h was true.
P(h) = Class prior probability.
The probability of hypothesis h being true (irrespective of the data)
P(d) = Predictor prior probability.

Probability of the data (irrespective of the hypothesis)
Naive Bayes is an easy and fast way to predict data class. Using this, multi-level forecasting can be done. When the independence assumption is valid, Naive Bayes is much more powerful than other algorithms such as logistic regression. Also, you will need less training data. However, Naive Bayes suffers from the following drawbacks:

- If a category variable belongs to a category that is not tracked in the training set, the model gives it a probability of 0, which prevents prediction.
- Naive Bayes assumes independence between its features. In real life, it is difficult to collect data that contains completely independent features [92-93-94-95-96-97-98-99-100-101-102].

$$y = \arg\max_{c_j \in C} \sum_{h_i \in H} P(c_j | h_i) P(T | h_i) P(h_i) \qquad (9)$$

here y is the predicted class, C is the set of all possible classes, H is the hypothesis space, P refers to a probability, and T is the training data. As an ensemble, the Bayes optimal classifier represents a hypothesis that is not necessarily in H. The hypothesis represented by the Bayes optimal classifier, however, is the optimal hypothesis in ensemble space (the space of all possible ensembles consisting only of hypotheses in H).

### 4.1.2. K Nearest Neighbors

A pattern recognition algorithm that uses the training dataset to find the nearest relatives in the following examples. When k-NN is used in classification, you calculate to place the data in the nearest neighbor group. If K = 1, it is in the nearest 1 class. K is classified by a multi-neighborly survey. The Neighbors K-Nearest algorithm uses the entire data set as a training set instead of dividing the data set into a training set and a test set.

The most intuitive nearest neighbour type classifier is the one nearest neighbour classifier that assigns a point *x* to the class of its closest neighbour in the feature space, that is:

$$C_n^{\ln n}(x) = Y(1) \qquad (10)$$

When a result is required for a new data sample, the KNN algorithm passes through the entire data set to find the k-samples closest to the new sample or the number of k samples similar to the new record and then subtracts the average (for a regression problem) or Mode (maximum class) for a classification problem. The value of k is specified by the user. The similarity between the samples is calculated using criteria such as Euclidean distance and Hamming distance [103-104-105-106-107-108-109-110].

### 5.1.2. Decision Tree

The decision tree is a supervised learning algorithm that is suitable for classification problems because it can order classes at an accurate level. It works like a flowchart, dividing data points from "tree trunk" to "branch" to "leaf" at the same time into two similar clusters, where the clusters become quite similar. Creates these classifications into categories and allows organic classification to be performed under limited human supervision [111-112-113-114-115-116].

A decision tree consists of *split nodes* $N^{Split}$ and *leaf nodes* $N^{Leaf}$. An example is illustrated in Fig. 4 Each split node $s \in N^{Split}$ performs a split decision and routes a data sample *x* to the left child node cl(s) or to the right child node cr(s). When using axis-aligned split decisions the split rule is based on a single split feature f(s) and a threshold value θ(s):

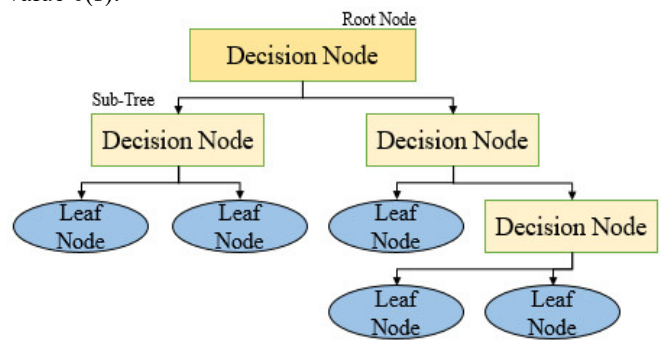

**Fig. 4**. Structure of the decision tree algorithm

Figure 4. An example of a decision tree that predicts whether or not to go hiking today. Split nodes (green) evaluate the data and route to the next node. Leaf nodes (blue) contain the possible outputs of the decision tree. Starting at the root node, the data is routed through the tree based on the split rules. Finally, a leaf node is reached which contains the decision output. In this example the output is "yes" or "no".

### 6.1.2. Random Forest

The random forest algorithm is a decision tree extension, in which you first create decision trees on some axes with instructional data, then place your new data as a "random forest" in one of the trees. Random forests or random forests are a set learning method for classification, regression, and other tasks. To build a random forest, you need to train a large number of decision trees on random samples of training data. Random forest output has the highest results among individual trees. Random Forests Due to the nature of the algorithm, they were able to successfully equip the equipment. Basically, it connects your data on average to connect it to the nearest tree on the data scale. Stochastic forest models are useful because they solve the problem of deciding to "force" data into an unnecessary group [117-118-119-120-121].

The training algorithm for random forests applies the general technique of bootstrap aggregating, or bagging, to tree learners. Given a training set $X = x_1, ..., x_n$ with responses $Y = y_1, ..., y_n$, bagging repeatedly ($B$ times) selects a random sample with replacement of the training set and fits trees to these samples:

$$For\ b = 1, ..., B:$$

1. Sample, with replacement, $n$ training examples from $X, Y$; call these $X_b, Y_b$.
2. Train a classification or regression tree $f_b$ on $X_b, Y_b$.

After training, predictions for unseen samples $x'$ can be made by averaging the predictions from all the individual regression trees on $x'$:

$$f = \frac{1}{B} \sum_{b=1}^{B} f_b(x') \qquad (11)$$

### 7.1.2. Support vector machine

Support vector machine is one of the supervised learning methods that is used for classification and regression. This algorithm uses algorithms to train and classify data in degrees of polarity and bring them to a higher degree of X / Y prediction. For a simple visual explanation, SVM uses a technique called kernel trick to convert your data, and then, based on that conversion, finds the optimal boundary between possible outputs. Simply performs very complex conversions, then specifies how to separate your data based on the tags or outputs you define [122-123-124-125-126-127]. For a nonlinear SVM, the cost function can be computed as:

$$\min_{0} C \sum_{i=1}^{m} \left( y_i c_1(\theta^t x_i) + (1-y_i) c_0(\theta^T x_i) \right) \qquad (12)$$

$\theta^T x \geq \phi, \theta^T x \geq \phi - 1.$

The SVM classifier is used when the dataset is less. SVM is not well suited to bigger datasets. For image classifications, CNN is the best approach to be used in place of the SVM (Figure 5).

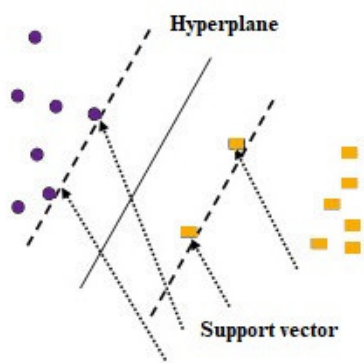

**Fig. 5**. Structure of the support vector machine

Define an optimal hyperplane: maximize margin Extend the above definition for non-linearly separable problems: have a penalty term for misclassifications.

Map data to high dimensional space where it is easier to classify with linear decision surfaces: reformulate problem so that data is mapped implicitly to this space. To define an optimal hyper plane we need to maximize the width of the margin (w).

### 8.1.2. CART Algorithm

Classification and Regression Trees (CART) is one of the implementations of decision trees. Non-terminal nodes of classification and regression trees are root node and internal node. The end nodes are the leaf nodes. Each non-terminal node represents a single input variable (x) and a divisor point on it. Leaf nodes represent the output variable (y). To predict, the model is used as follows: Follow the three divisions to reach the leaf node and generate the value in the leaf node [128-129-130-131]. In data mining, decision trees can be described also as the combination of mathematical and computational techniques to aid the description, categorization and generalization of a given set of data.

Data comes in records of the form:

$$(x, Y) = (X_1, X_2, X_3, ..., X_k, Y) \qquad (12)$$

The dependent variable, Y, is the target variable that we are trying to understand, classify or generalize. The vector X is composed of the features, $X_1, X_2, X_3$ etc., that are used for that task.

**Gini impurity:** Used by the CART (classification and regression tree) algorithm for classification trees, Gini impurity is a measure of how often a randomly chosen element from the set would be incorrectly labeled if it was randomly labeled according to the distribution of labels in the subset.

**Information gain:** Used by the ID3, C4.5 and C5.0 tree-generation algorithms. Information gain is based on the concept of entropy and information content from information theory.

Entropy is defined as below:

$$H(T) = I_E(p_1, p_2, ..., p_j) = \sum_{i=1}^{j} {}_i p_i \log_2 p_i \qquad (13)$$

**2.2. Ensemble learning techniques**

Ensemble learning is a problem-solving method by building several ML models and combining them. Ensemble learning is primarily used to improve the performance of classification, forecasting, and performance approximation models. Other groups learning programs include model decision review, optimal feature selection for construction models, incremental learning, and unstable learning. [132-133-134-135] Below are some common Ensemble learning algorithms.

### 1.2.2 Bagging

stands for Bootstrap collection. This is one of the oldest Ensemble learning algorithms that work very well. To ensure diversity of classifications, use bootstrap versions of training data. This means that different subsets of educational data are randomly drawn from the educational data set by substitution. Each subset of training data is used to teach different classifications of the same type. Then, separate classifiers can be combined. To do this, you must make their decisions by a simple majority of votes. The class determined by most classifiers is the decision of the group [136-137-138-139-140]. Mathematically, Bagging is represented by the following formula,

$$\widehat{f_{bag}} = \widehat{f_1}(x) + \widehat{f_2}(x) + ... + \widehat{f_b}(x) \qquad (14)$$

The term on the left hand side is the bagged prediction, and terms on the right hand side are the individual learners.

### 2.2.2 Boosting

This group of group algorithms is similar to bagging. Boosting also uses different classifications to re-sample the data and then selects the optimal version by a majority vote. In reinforcement, you regularly train weak classifiers to bring them together in a strong classification. When classifications are added, they are usually attributed at some point to describe the accuracy of their prediction. Weights are recalculated after poor classification is added to the group. Wrongly classified inputs gain more weight and incorrectly classified samples lose weight; therefore, the system focuses more on samples that have been misdiagnosed [140-141-42-143-144-145]. The following illustration gives a visual insight into the boosting algorithm.

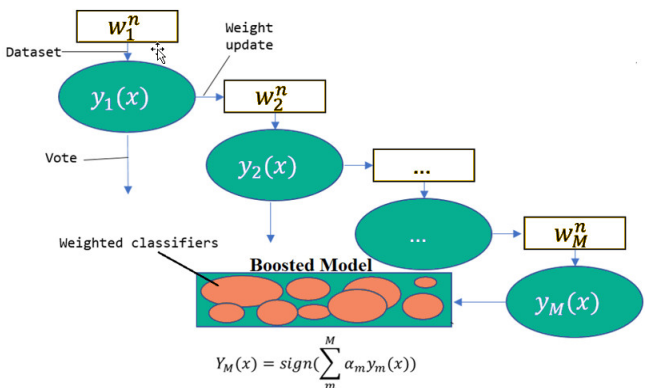

**Fig. 6.** Structure of the Boosting

Here the different base classifiers are each build on a weighted dataset where the weights of the single instances in the dataset depend on the results the previous base classifiers had made for these instances. If they have misclassified a instance, the weight for this instance will be increased in the next model while if the classification was correct, the weight remains unaltered. The final decision making is achieved by a weighted vote of the base classifiers where the weights are determined depending on the misclassification rates of the models. If a model has had a high classification accuracy, it will get a high weight while it gets a low weight if it has had a poor classification accuracy. Bootstrap ping (Source: https://www.python-course.eu/Boosting.php).

### 3.2.2 AdaBoost

Adaptive Boosting is a meta-algorithm for machine learning formulated by Yoav Freund and Robert Schapire, who won the 2003 Gödel Prize for their work. It can be used in conjunction with many other types of learning algorithms to improve performance [161]. AdaBoost refers to a particular method of training a boosted classifier. A boost classifier is a classifier in the form:

$$F_T(x) = \sum_{t=1}^{T} f_t(X) \qquad (15)$$

where each $F_t$ is a weak learner that takes an object X as input and returns a value indicating the class of the object. For example, in the two-class problem, the sign of the weak learner output identifies the predicted object class and the absolute value gives the confidence in that classification. Similarly, the T classifier is positive if the sample is in a positive class and negative otherwise.

### 4.2.2 Stacking

A stacking is a collection learning technique that combines several classifications or regression models through a meta-classifier or an organizer. Basic level models are trained based on a complete set of training, then the meta-model is taught as a feature in the output of basic level models [145-145-147-148-149-150-151].

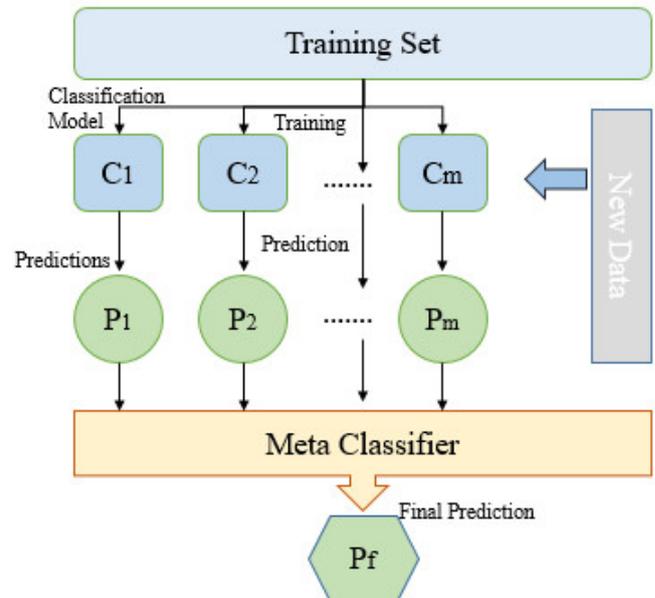

**Fig. 7.** Structure of the Staking algorithm

The algorithm can be summarized as follows (source: [162-163]):

**Algorithm Stacking**
**Input**: training data D = {Xi, yi}$^m_{i=1}$
**Ouput**: ensemble classifier **H**
Step 1: learn base-level classifiers
for t=1 to T **do**
    learn $h_t$ based on D
**end for**
**Step 2**: construct new data set of predictions
for i=1 to m **do**
    $D_h$ = {Xi, yi}, where Xi } = {h1 (Xi), ..., $h_T$(Xi)}
**end for**
**Step 3**: learn a meta-classifier
learn **H** based on $D_h$
return **H**

### 5.2.2 Neural networks

The neural network is a sequence of neurons that are connected by synapses and recall the structure of the human brain. However, the human brain is even more complex. What is very interesting about neural networks is that they can be used for almost anything from spam filtering to computer vision. However, they are commonly used for machine translation, anomaly detection and risk management, speech recognition and language production face recognition, and more. Figure 8 shows the structure of the neural network algorithm.

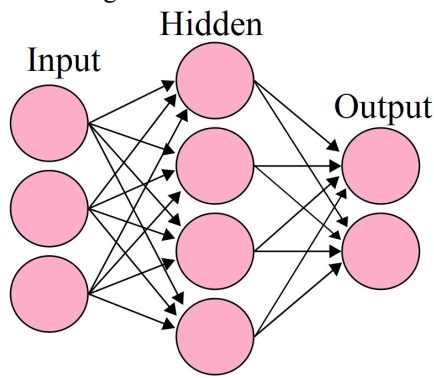

**Fig. 8.** Structure of the neural network

Here, each circular node represents an artificial neuron and an arrow represents a connection from the output of one artificial neuron to the input of another.

**Each neuron processes signals in the same way**; but how can a different result be obtained? The synapses that connect the neurons are responsible for this. Each neuron can have many synapses that weaken or amplify the signal. Also, neurons can change their properties over time. By selecting the correct synapse parameters, we will be able to obtain the correct results of converting input information to output.
There are different types of NN: Feedforward (FFNN) neural networks and receivers are very simple, there are no loops or cycles in the network. In practice, such networks are seldom used, but they are often combined with other types to obtain new networks.

#### 6.2.2 The Hopfield Network (HN)

HN is a neural network perfectly connected to the symmetric matrix of links. Such networks are often called corporate memory networks. Just like a person who can imagine the second half of a table by imagining it, this network returns it to its full state by receiving a noisy table.

#### 7.2.2 Convolution neural networks

CNNs and deep convolution neural networks (DCNNs) are very different from other types of networks. They are commonly used for image processing, audio, or video tasks. A common way to use CNN is to classify images. It is interesting to observe different types of neural networks [152-153-154-155-156-157-158-159-160].
So here are the top 10 data from the list of data mining algorithms. We hope this article is a little clearer based on these algorithms. This study examines the prognosis of chronic diseases using machine learning techniques along with other Ensemble learning methods. In this study, the three data sets mentioned in Section 3 will be used and the algorithms examined in Section 2 will be used to implement this data set. The performance of the models is evaluated using the common confusion matrix and ROC performance analysis. The final model of readmission is based on the positive rate, sensitivity, and actual characteristics based on the best model.

### 3. Material and Methods

1.3. Dataset
1.1.3. Diabetes Database
Diabetes data set used in this study of diabetic walk patients from the UCI site at (HTTP://archive.ics.uci.Edu/ml/datasets/diabetes) that this data set has 9 variables and 768 records that these variables and abbreviations are given in Table 1.

**Table 1**
Problem variables and acronyms

| | |
|---|---|
| P.NO | Number of times pregant |
| PG | Plasma Glucose Concentration |
| DBP | Diastolic Blood Pressure |
| TSFT | Triceps Skin Fold Thickness |
| SI | Two Hour Serum Insulin |
| BMI | Body moss index |
| DPF | Diabetes Pedigree Function |
| AGE | Age |
| C | Class variable |

2.1.3. Cancer dataset
The Breast Cancer Patient Database from the UCI website contains 32 variables and 569 records in the University of California, USA machine learning database (HTTPS://archive.ics.uci.Edu/ml/datasets / Breast + Cancer + Wisconsin + (diagnostic)). This database includes risk factors for bulk thickness, cell size uniformity, cell shape uniformity, edge adhesion, epithelial tissue cell volume, naked nuclei, chromatin long, normal nucleus, and cell division, collected in Wisconsin, USA.

**Table 2**
Breast Cancer Database

| Dataset | No. of attribute | No. of instances | No. of class |
|---|---|---|---|
| Wisconsin diagnosis of breast cancer (WDBC) | 32 | 569 | 2 |

3.1.3. **Heart Database**:
The data set of heart patients available in HTTPS://archive.ics.uci.Edu/ml/datasets/statlog+(heart) has been used. This data set has 13 useful variables and 270 records, which these variables and abbreviations in Tables 3 are given.
**Table 3**
Heart disease data set

| ID | Meaning | Type |
|---|---|---|
| Age (age in the year) | Patient age | Integer |
| sex | Gender | Numerical two values |
| chest pain | The location of chest pain | Numeric values |
| blood pressure | blood pressure | Integer |
| cholesterol | Cholesterol content | Integer |
| blood sugar | Blood sugar | Numerical two values |
| electrocardiographic | ECG result (electrocardiographic) | Three-digit number (0, 1, and 2) |
| heart rate | heartbeat | Integer |
| exercise-induced | Angina result from an exercise test | Numerical two values |
| depression | ST depression rate | Real number |
| slope | ST Exercise Test Result | Numeric values |
| ca | Is there a blockage in the arteries or not? | Numerical two values |
| Thal | Thalassemia problem | Numeric values |
| C(Objective variable) | Is the patient at risk for a heart attack? | Two-digit number |

2.3. How to work hybrid neural network-based stacking

Because the use of an intelligent machine learning algorithm is not effective and accurate in diagnosing and predicting diseases and has not been successful in many scenarios alone, in this study we use Ensemble learning to diagnose chronic diseases (diabetes, heart, and cancer). The purpose of this paper is to improve the accuracy and speed of diagnosis of chronic diseases in the context of intelligent networks and compare basic, meta, and hybrid algorithms on diseases, and build a meta-hybrid algorithm using the NN algorithm for accurate diagnosis of diseases. NN algorithms are new algorithms.

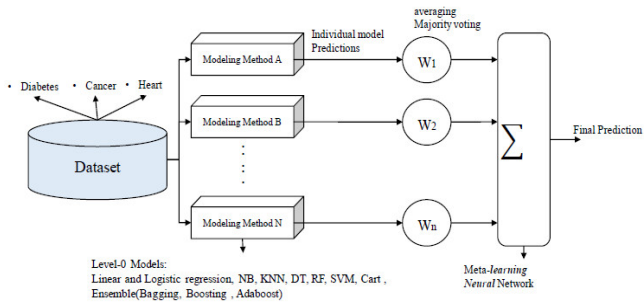

**Fig. 9.** Schematic of the model

. In this paper, we want to use this algorithm as a meta-hybrid based and the top 10 machine learning algorithms as the basic algorithms in the stack generalization algorithm to predict chronic diseases and implement a hybrid meta-algorithm for forecasting. The use of a new meta-learner (DeepNN_SG) in the stacking learning method and the top 10 algorithms as a basis are some of the innovations of this paper. Fig 9 shows a schematic and Figure 10 shows the proposed flowchart.

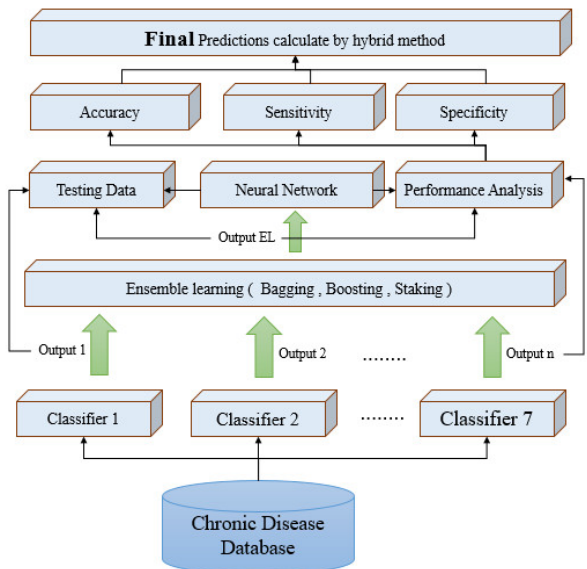

**Fig. 10.** Flowchart of the model used

### 3.3. Evaluation Metrics

To evaluate the methods used and determines how it was possible to consider among the existing models, the model that has the highest accuracy of prediction compared to other methods, which is compared to the batch methods. Configuration and Finding Appropriate and Efficient Method in this article, we have used cost-benefit analysis (disintegration matrix), ROC curve, and other issues related to model selection such as accuracy, etc.

Performance measurement is used to determine the effectiveness of the classification algorithm in such a way that in two-category classification problems, the classification cost can be represented by a cost matrix, thus making a mistake for two types of positive errors. (FP) And false-negative (FN) and two types of classifications true positive (TN) and true negative (TN) that have different costs and benefits, which are shown in Table 4 [9 - 10].

**Table 4**
Confusion matrix

| Confusion Matrix | | **Classified As:** | |
|---|---|---|---|
| | | Negative | **Positive** |
| Actual | Negative | TN | **FP** |
| Class | Positive | FN | **TP** |

- **True Positive (TP)** are positive items that are correctly classified as positive.
- **Real negatives (TN)** are negatives that are correctly identified as negatives.
- **False positives (FPs)** are negatives that are classified as positive (mistakenly, someone who is not ill has breast cancer.) are more expensive.
- **False negatives (FNs)** are positives that are negative (for example, misdiagnosing the breast cancer, patient does not mean breast cancer).
- **Classification accuracy** is the simplest measure of performance accuracy, which means the percentage of correctly predicted batches and is calculated from the following formula.

$$Accuracy = \frac{TP+TN}{TP+FP+TN+FN} \quad (16)$$

- **Sensitivity**: Real positive rate: If the result is positive for the person, in a few percent of cases, the model will be positive, which is calculated from the following formula.

$$Sensitivity = \frac{TP}{TP+FN} \quad (17)$$

**Properties**: Real negative rate: If the result is negative for the person, in a few percent of cases, the model will also be a negative result, which is calculated from the following formula.

$$Specificity = \frac{TN}{TN+FP} \quad (18)$$

- **Receptor Characteristics (ROC):** which, for evaluating models, draws a graph, each of which has a higher graph. The model is a better model that uses the following criteria to compare models. Categorization is used graphically.

$$TPR = \frac{TP}{TP+FN} \qquad (19)$$

$$FPR = \frac{FP}{FP+TN} \qquad (20)$$

- **PPV**: If the model is positive, how likely is it that a person will develop breast cancer?

$$PPV = \frac{TP}{TP+FP} \qquad (21)$$

- **NPV**: If the model is positive, how likely is it that a person will develop breast cancer?

$$NPV = \frac{TN}{TN+FN} \qquad (22)$$

## 4. Results

Classification are one of the most important aspects of supervised learning. In this paper, we discuss various classification algorithms such as logistic regression, decision tree, random forests, and many more in diagnosing chronic diseases. We used the classification features of the algorithm and how they worked.

**Table 5**
Outcomes of Diabetes

|  | Specificity | Sensitivity | Accuracy |
|---|---|---|---|
| Logistic regression | 77 | 78 | 78.5 |
| Naïve Bayes | 75 | 73 | 75 |
| K-Nearest Neighbors | 78 | 80 | 81 |
| Decision Tree | 79 | 77 | 81.5 |
| Random Forest | 84 | 87 | 88 |
| Support Vector Machine | 64 | 63 | 65 |
| **CART** | 58 | 52 | 58 |

**Table 6**
Outcomes of heart disease

|  | Specificity | Sensitivity | Accuracy |
|---|---|---|---|
| Logistic regression | 68 | 59 | 68 |
| Naïve Bayes | 80 | 77 | 81 |
| K-Nearest Neighbors | 77 | 72 | 77 |
| Decision Tree | 73 | 70 | 73 |
| Random Forest | 83 | 84 | 84 |
| Support Vector Machine | 81 | 80 | 81 |
| CART | 76 | 77 | 78 |

**Table 7**
Outcomes of cancer

|  | Specificity | Sensitivity | Accuracy |
|---|---|---|---|
| Logistic regression | 84 | 80 | 84 |
| Naïve Bayes | 88 | 89 | 89 |
| K-Nearest Neighbors | 77 | 73 | 78 |
| Decision Tree | 75 | 78 | 79 |
| Random Forest | 77 | 83 | 88.5 |
| Support Vector Machine | 83 | 85 | 86 |
| CART | 53 | 55 | 56 |

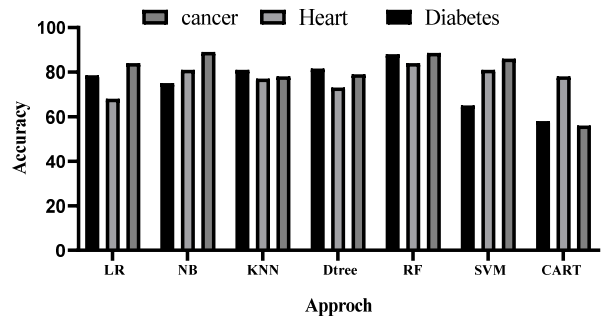

**Fig. 11.** Results of basic algorithms on chronic diseases

**Table 8**
Diabetes Outcomes

|  | Specificity | Sensitivity | Accuracy |
|---|---|---|---|
| Bagging | 88 | 89 | 89 |
| Boosting | 90 | 92 | 92 |
| Stacking | 97 | 95 | 98 |
| Deep NN | 98 | 100 | 99 |

**Table 9**
Outcomes of heart disease

|  | Specificity | Sensitivity | Accuracy |
|---|---|---|---|
| Bagging | 92 | 91 | 92 |
| Boosting | 93 | 92 | 94 |
| Stacking | 95 | 98 | 97 |
| Deep NN | 98 | 97 | 99 |

**Table 10**
Outcomes of cancer

|  | Specificity | Sensitivity | Accuracy |
|---|---|---|---|
| **Bagging** | 86 | 89 | 89 |
| **Boosting** | 88 | 86 | 88 |
| **Stacking** | 96 | 98 | 98 |
| **Deep NN** | 97.5 | 98 | 98.5 |

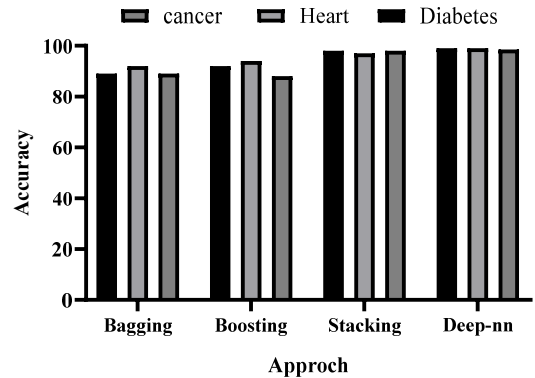

**Fig. 12.** Results of basic algorithms on chronic diseases

## 5. Discussion

Predicting the diagnosis of chronic diseases has always been an important challenge for researchers and physicians. Today, with the help of artificial intelligence, the possibility of solving these challenges by using previously recorded information from patients has been realized to a large extent. With low-cost hardware and software technologies, data are automatically stored with better quality and in higher volumes, and with better analysis, this huge amount of data is processed more efficiently and effectively. The main purpose of this article is to introduce several widely used and well-known data mining algorithms in breast cancer. In this article, we look at the different algorithms used to classify chronic diseases. We also analyzed their advantages and limitations in diagnosing chronic diseases.

The purpose of this paper was to provide a clear picture of each of the classification algorithms in the accurate diagnosis of chronic diseases. In this paper, classification algorithms were used to predict chronic diseases. Each algorithm was evaluated for a database obtained from a reputable UCI website. We compared their results with the results of a similar study in the form of tables and proposed the best possible algorithm as the result. During the performed processes, it was concluded that the group algorithm has higher predictive accuracy than the basic algorithms. The results of these algorithms not only help physicians make better decisions but also reveal some hidden and unknown patterns that may not have received much attention.

There is a wide variety of classifications used in this field. Many new algorithms have been developed and tested for the classification of chronic diseases. Now it is time to review the classification algorithms for classifying chronic diseases. Approach. We reviewed health care devices and the machine learning literature from 2007 to 2020 to identify new classification approaches that have been explored for the design of chronic diseases and to use them to provide a new approach to improving the diagnosis of chronic diseases.

## 6. Conclusion and future work

Chronic diseases are among the most costly and common health issues; therefore, trying to change the natural course of these diseases includes not only screening for early detection of diseases, but also special attention to prevention. The use of classification algorithms in the construction of decision-making devices can help physicians decide on the type of disease, and specialists can take action to treat patients according to the type of disease, during which the potential risks of not diagnosing the disease in a timely or correct manner will be reduced. In this article, the top 10 machine learning algorithms in predicting and diagnosing chronic diseases were introduced, and with the introduction of each algorithm, a background of research done on diseases using the algorithm, its results, and current research in this regard are presented. Ensemble learning algorithms, in various researches, have usually provided better and more accurate results in terms of accuracy, sensitivity and specificity. The success of these algorithms depends on several factors such as the presence of required variables, larger database, small number of missing data and access to accurate data [9-10].

**In the future research**, we plan to use artificial intelligence to diagnose neurological diseases.

**Conflict of interest**
The authors declare that no conflict of interest exists.

**Acknowledgment**
We are thankful to our colleagues who provided expertise that greatly assisted the research.

**Source of funding**
All the funding of this study was provided by the authors.